%% file: FA2025_template.tex
\documentclass[11pt]{article}
\usepackage{fa2025}
\usepackage{amsmath}
\usepackage{amsfonts}
\usepackage{cite}
\usepackage{url}
\usepackage{graphicx}
\usepackage{color}
\usepackage{siunitx}
\usepackage[utf8]{inputenc}

\title{Latent FxLMS: Accelerating Active Noise Control with Neural Adaptive Filters}

\multauthor
{Kanad Sarkar$^{1*}$ \hspace{1cm} Austin Lu$^1$ \hspace{1cm} Manan Mittal$^2$ } { \bfseries{Yongjie Zhuang$^{2,5}$ } \hspace{1cm} \bfseries{Ryan M. Corey$^{3,4}$ \hspace{1cm} Andrew C. Singer$^{1,2}$}\\
  $^1$ Electrical and Computer Engineering, University of Illinois Urbana-Champaign, USA\\
$^2$ Electrical and Computer Engineering, Stony Brook University, USA\\
$^3$ Electrical and Computer Engineering, University of Illinois Chicago, USA\\
$^4$ Discovery Partners Institute, University of Illinois System, USA\\
$^5$ Amazon Web Services$^\dagger$
\correspondingauthor{kanads2@illinois.edu
\\
$^\dagger$This work does not relate to the author’s position at Amazon}{Sarkar et al.}
}

\begin{document}

\maketitle
\begin{abstract}
Filtered-X LMS (FxLMS) is commonly used for active noise control (ANC), wherein the soundfield is minimized at a desired location. Given prior knowledge of the spatial region of the noise or control sources, we could improve FxLMS by adapting along the low-dimensional manifold of possible adaptive filter weights. We train an auto-encoder on the filter coefficients of the steady-state adaptive filter for each primary source location sampled from a given spatial region and constrain the weights of the adaptive filter to be the output of the decoder for a given state of latent variables. Then, we perform updates in the latent space and use the decoder to generate the cancellation filter. We evaluate how various neural network constraints and normalization techniques impact the convergence speed and steady-state mean squared error. Under certain conditions, our Latent FxLMS model converges in fewer steps with comparable steady-state error to the standard FxLMS.
\end{abstract}

\input{text/Intro}

\input{text/basics}

\input{text/NeuralModels}

\input{text/Experiments}

\input{text/Conclusion}

\section{Acknowledgments}
The authors acknowledge support from the Foxconn Interconnect Technology sponsored Center
for Networked Intelligent Components and Environments (C-NICE) at the University of Illinois at
Urbana-Champaign.

\bibliography{fa2025_template}

%
%
%

\end{document}

%% file: text/Intro.tex
\begin{figure}[ht]
 \centerline{
 \includegraphics[width=0.8\columnwidth]{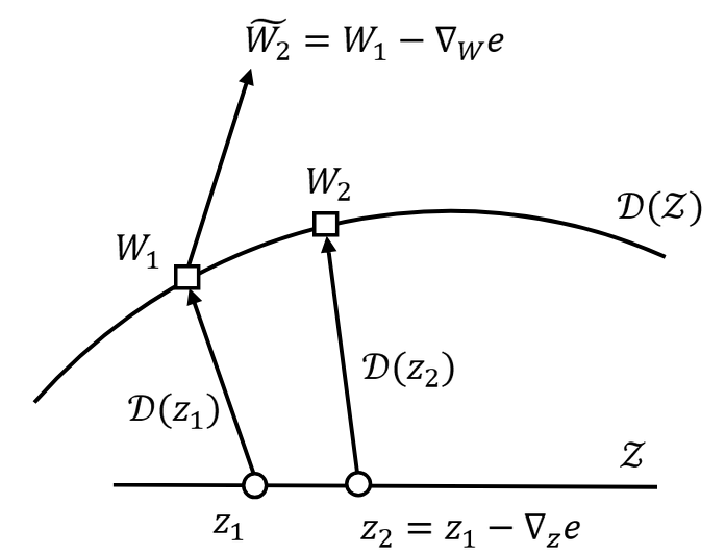}}
 
 \caption{ Diagram depicting the core element of Latent FxLMS.  Instead of adapting the weights directly, a low-dimensional representation is updated based on the error gradient. The new weights are determined by the decoder's output at the updated latent representation. This method allows for faster convergence by constraining the updated weights to lie on the manifold of learned filters.
 }
 \label{fig:pic}
\end{figure}
\section{Introduction}\label{sec:introduction}

Active Noise Control (ANC) uses loudspeakers to suppress the effect of an undesirable noise source for a listener at a given location.  A standard ANC algorithm is the Filtered-x Least Mean Squares (FxLMS) algorithm, which adaptively filters a reference measurement of the noise signal to form a cancellation signal output by the control speakers. This is an effort to minimize the energy from the noise source at the desired location, where an error microphone is ideally placed. The ``Filtered-x'' component filters the reference signal with the impulse response between the control speakers and the error microphone, which is then used in the adaptive filter update.  A review of ANC methods can be found in \cite{survey1}.


While it is well known that the filter adapts to time-varying noise statistics \cite{survey1}, it can also adapt to time-varying noise source positions when the noise source is farther from the error microphone than the control speaker \footnote{ If the primary source is closer to the microphone compared to the secondary source, knowledge of future samples of the noise signal is required to cancel sound at the microphone. This goes against the requirement that the adaptive filter is causal. }.   As the noise source moves within a spatial region, the adaptive filter will change within a subset of $\mathbb{R}^L$, where $L$ is the filter length. When a manifold of a lower dimension than $L$ can approximate the space of converged adaptive filters, a low-dimensional representation of this space can be exploited to improve the convergence of the filter adaptation.  When the possible weights are constrained to a known Riemannian manifold, it can be shown that geometrically constrained adaptive filtering algorithms converge faster to a lower steady-state error than the standard algorithms \cite{Riemann}. However, obtaining a closed-form representation of the motion-based manifold of filter coefficients is difficult.

The proposed Latent FxLMS (LFxLMS) attempts to spatially constrain the adaptive ANC filter by combining the FxLMS algorithm and recent work on neural-network-based system identification (NNSI). The NNSI framework constrains the FIR filter to be the output of a decoder of a pre-trained auto-encoder, and adaptations are performed within the latent space of the decoder. Previous work on acoustic NNSI uses various approaches to track time-varying impulse responses \cite{Karim, EKF}.  A topologically aware variational auto-encoder was used to speed up the convergence of RLS \cite{Karim}. Furthermore, the authors in \cite{EKF} applied this framework to an extended Kalman filter to improve acoustic echo cancellation performance. The model developed in \cite{metaAF} is similar to our work in that it uses a neural network to provide gradient updates to an adaptive filter in a real-time acoustic system. However, \cite{metaAF} achieves this through a meta-learning framework, where their model output gives the weight gradient, while NNSI provides a gradient in the latent space. This paper also presents novel methods for normalized latent updates and examines their impact on adaptive filter convergence.

The LFxLMS algorithm involves training an auto-encoder on a dataset of converged FxLMS adaptive filters, each corresponding to a different noise source location sampled from a region in space. A wide variety of existing literature on impulse response auto-encoder models can be incorporated into this work. Models such as \cite{naf, brendel} used physics-based and signal-based constraints to interpolate acoustic filter parameters. This paper studies two neural constraints that can be used alongside acoustic-based constraints in future work. The first constraint is the requirement of a disentangled latent space through variational auto-encoders (VAE) \cite{vae_paper, infovae}. The second constraint requires the convex combination of samples in the impulse response dataset to be reconstructed with the auto-encoder \cite{mixup}. 

To our knowledge, this is a novel combination of neural networks and ANC. Early ANC work directly used neural networks as adaptive filters, replacing the adaptive FIR filter with shallow artificial neural networks \cite{survey2, flann, neuraliir}. With advances in machine learning since 2010, progressively deeper neural networks have been employed. These models have been trained on a variety of noise statistics, such as those of hums or impulsive sounds, and either provided gradient updates to the filter or completely replaced the adaptive filter based on the noise source statistics \cite{dlanc, rlanc,rlanc2,mcmlanc}.  While changing noise statistics are well accounted for, these works do not consider dynamic environments that change \textit{during} ANC, e.g., due to the changing position of the noise source. 



This paper provides evidence that when the noise source is spatially constrained to be in a region, the LFxLMS algorithm will converge faster than the traditional FxLMS. The impact of neural-network constraints and normalized latent gradients are also studied.

%% file: text/basics.tex
\section{FxLMS and Latent FxLMS}\label{sec:background}

This paper assumes a-priori knowledge of the Nyquist sampled noise signal, which we denote as $x = [x_0, x_1, \dots]$. Assume all filters have the same filter length $L$. Let the primary noise acoustic path $\mathbf{p} = [p_0, \dots, p_{L-1}]^T$ be the impulse response between the noise source and the error micro; hone. Let the control or secondary path $\mathbf{g} = [g_0, \dots, g_{L-1}]^T$ be the impulse response between the control speaker and the error microphone, and assume there is an estimate of the secondary path denoted as $\mathbf{\hat{g}}$. Furthermore, let $\mathbf{w}[n] = [w_0, \dots, w_{L-1}]^T$ be the adaptive filter coefficients. Finally, let $e = [e_0,e_1, \dots]$ denote the sampled signal the error microphone receives. In this discrete-time, linear, and time-varying scenario, we want to solve the following minimization problem,
\begin{equation}\label{optim}
    \begin{aligned}
   \mathbf{w}_{\text{opt}}&=\arg \min_\mathbf{w} E_x[(e)^2 ],   
\end{aligned}
\end{equation}
where $E_x[\cdot]$ is the expectation over the input signal $x$, and the microphone signal $e$ is given by
\begin{equation}
e = \mathbf{p}*x + \mathbf{g}*\mathbf{w}*x ,
\end{equation}
where $*$ is the linear convolution operator.


\subsection{Filtered-X LMS algorithm}
\begin{figure}[ht]
 \centerline{
 \includegraphics[width=7.8cm]{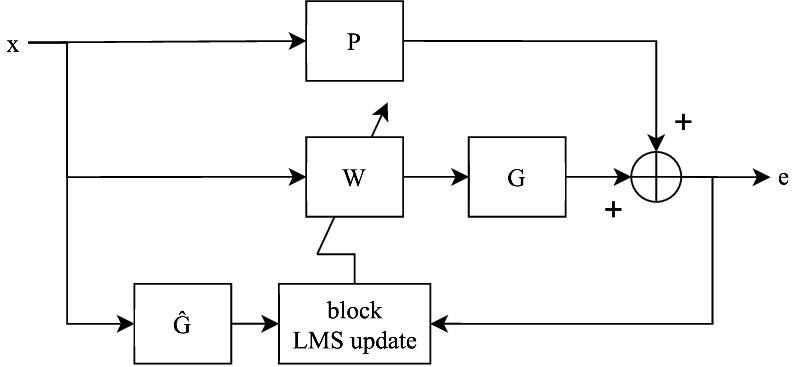}}
 \caption{ The FxLMS algorithm layout. The goal of ANC is to acoustically cancel the image of the noise source $\mathbf{p}*x$ at a desired location. A control speaker emits a filtered reference of the noise signal, whose image $\mathbf{g}*\mathbf{w}*x$ negates the noise source.}
 \label{fig:fxlms}
\end{figure}

The "Filtered-X" component of the FxLMS refers to filtering $x$ with the estimated secondary path $\mathbf{\hat{g}}$ to form the filtered signal $\hat{x} = \mathbf{\hat{g}} * x$ whose samples are given by
\begin{equation}
\begin{aligned}
    \hat{x}_n & = \mathbf{\hat{g}}^T \mathbf{x}_n \\
    &= \sum_{i=0}^{L-1} \hat{g}_i x_{n-i},
\end{aligned}
\end{equation}
where $\mathbf{x}_n = \begin{bmatrix}
    x_n & x_{n-1} & \cdots & x_{n-L+1}
\end{bmatrix}^T$. This paper uses the normalized update of the FxLMS algorithm, given as
\begin{equation}
\begin{aligned}
\mathbf{w}[n] = \mathbf{w}[n-1] - \mu e_n \frac{\mathbf{\hat{x}}_n}{\epsilon + \|\mathbf{\hat{x}}_n\|^2},
\end{aligned}
\end{equation}
where $\mu$ is the user-selected step size, and $\mathbf{\hat{x}_n} = [\hat{x}_n, \dots, \hat{x}_{n-L+1}] $, and $\epsilon >0$ is a small constant to prevent a division by zero. While the normalized FxLMS requires more computation, it is preferred under time-varying conditions \cite{Sayed}. For computational efficiency, we perform the adaptive filtering in blocks of samples, which means we average the weight update across samples before updating the filter.  A diagram of this algorithm is provided in Figure \ref{fig:fxlms}.

\subsection{Latent Filtered-x Least Mean Squared} \label{sec:lfxlms}

\begin{figure}[ht]
 \centerline{
 \includegraphics[width=7.8cm]{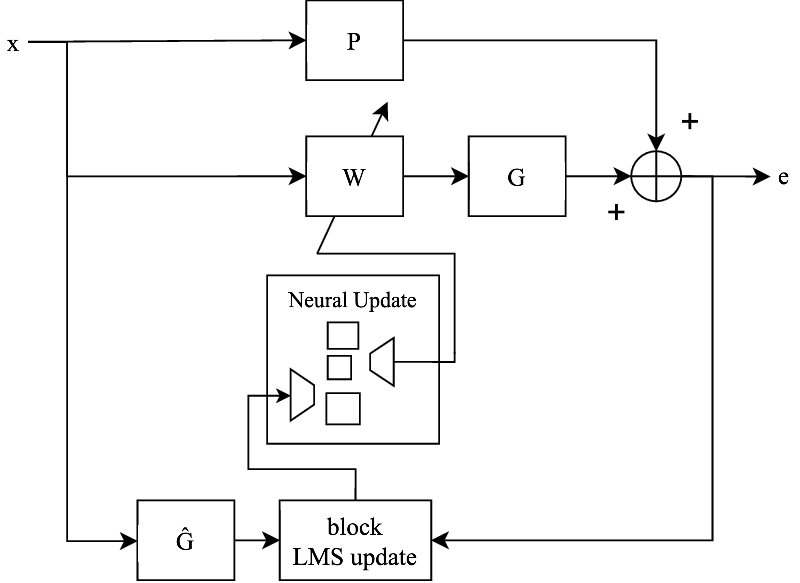}}
 \caption{ The latent FxLMS algorithm layout, where adaptations occur in the latent space of the decoder. The Neural Update block is expanded in Figure \ref{fig:neuralupdate}.}
 \label{fig:latentfxlms}
\end{figure}

\begin{figure}[ht]
 \centerline{
 \includegraphics[width=\columnwidth]{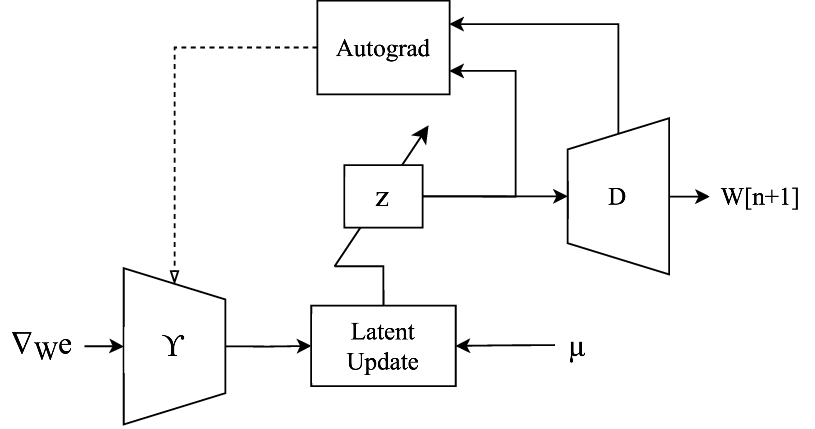}}
 
 \caption{  This model defines the latent variable update block, which inputs the weight gradient and outputs the next adaptive filter coefficients. The weight gradient is the input of $\Upsilon$, which is the Jacobian of the decoder at $z$.}
 \label{fig:neuralupdate}
\end{figure}

The Latent Filtered-x LMS (LFxLMS) algorithm assumes that the primary source position lies within a known room region and that we know the static secondary path. Thus, the space of possible adaptive filters $\mathcal{W}$ can be approximated by a manifold of a lower dimension than $L$. Under this manifold assumption, say that there exists a differentiable function $\mathcal{D}$ that maps a low-dimensional space $\mathcal{Z}$ to $\mathcal{W} \subset \mathbb{R}^L$. The LFxLMS algorithm requires $\mathbf{w}[n]$ to be in the range space of $\mathcal{D}$. 
\begin{equation}
    \mathbf{w}[n] = \mathcal{D}(\mathbf{z}[n])
\end{equation}
The optimization in Equation (\ref{optim}) is then reformulated as,
\begin{equation}
    \begin{aligned}
  \mathbf{z}_{\text{opt}}&=\arg \min_\mathbf{z} E_x[(e)^2 ]\\
   &= \arg \min_\mathbf{z} E_x[(\mathbf{p}*x + \mathbf{g}*\mathcal{D}(\mathbf{z})*x)^2].
\end{aligned}
\end{equation}
Assuming $\mathcal{D}$ is differentiable, we can adapt the latent variable using the Jacobian $\Upsilon$ evaluated at the latent variable,
\begin{equation}
    \Upsilon(\mathbf{z}^*) =\begin{bmatrix}
        \frac{\partial w_1}{\partial z_1}|_{\mathbf{z} = \mathbf{z}^*} & \dots & \frac{\partial w_L}{\partial z_1}|_{\mathbf{z} = \mathbf{z}^*}\\
        \vdots& \ddots & \vdots\\
        \frac{\partial w_1}{\partial z_k}|_{\mathbf{z} = \mathbf{z}^*} & \dots & \frac{\partial w_L}{\partial z_k}|_{\mathbf{z} = \mathbf{z}^*}
    \end{bmatrix},
\end{equation}
where $w_1,\dots, w_L$ are the filter coefficients. The Jacobian matrix $\Upsilon(\mathbf{z})$ linearly maps a tangent vector at $\mathbf{w}[n] =\mathcal{D}(\mathbf{z}[n])$ to a tangent vector at the latent vector $\mathbf{z}[n]$. Thus, we can map the FxLMS gradient to the corresponding gradient direction in the latent space. The chain rule yields the following Latent FxLMS update rule,
\begin{equation}
    \mathbf{z}[n] = \mathbf{z}[n-1] - \mu_z \Upsilon(\mathbf{z}[n-1]) (e_n \mathbf{\hat{x}}_n),
\end{equation}
where $\mu_z$ is the step-size in latent space. Figure \ref{fig:latentfxlms} is a diagram that describes the latent FxLMS algorithm, with Figure \ref{fig:neuralupdate} defining the latent update block. 

\subsection{Normalization Schemes for LFxLMS}

Using a normalized step size can improve the convergence of adaptive filters. Two possible approaches to normalization exist when adapting in $\mathcal{Z}$. 

\subsubsection{Data-normalized Latent LMS} \label{subsubdata}
The first approach treats the normalized gradient of the normal FxLMS as the input gradient into $\Upsilon(\mathbf{z})$,
\begin{equation}
    \mathbf{z}[n] = \mathbf{z}[n-1] - \frac{\mu_z}{\|\mathbf{\hat{x}}_n\|^2 + \epsilon} \Upsilon(\mathbf{z}[n-1]) (e_n \mathbf{\hat{x}}_n).
\end{equation}

\subsubsection{Latent-normalized Latent LMS}\label{subsublat}
The second approach involves normalizing the latent gradient,
\begin{equation}
\begin{aligned}
    \mathbf{z}[n] =  \mathbf{z}[n-1] - \frac{\mu_z \Upsilon( \mathbf{z}[n-1]) (e_n  \mathbf{\hat{x}}_n)}{\|\Upsilon( \mathbf{z}[n-1]) \mathbf{\hat{x}}_n\|^2 + \epsilon}.
\end{aligned}
\end{equation}

The data-normalized LFxLMS normalizes the weight gradient before being input $\Upsilon$, while the latent-normalized FxLMS normalizes the latent gradient. As shown in Section \ref{sec:experiments}, the normalization schemes can affect the convergence rate of the LFxLMS algorithm.

%% file: text/NeuralModels.tex
\section{Auto-Encoder models}
\label{sec:neural}

The NNSI framework requires an auto-encoder trained on the space of possible filter coefficients. An auto-encoder is the composition of an encoder $\mathcal{E}$, which maps $\mathcal{W} \subset \mathbb{R}^L$ to a lower-dimensional latent space $\mathcal{Z} \subset \mathbb{R}^k$, with a decoder $\mathcal{D}$, which maps the latent variable to an element of the input space. The LFxLMS algorithm constrains the weights to be in the range space of the decoder $\mathcal{D}$. Thus, LFxLMS performance is heavily influenced by how the auto-encoder is trained.

This paper uses the same structure for all models presented to verify the LFxLMS performance using various neural-based constraints. The encoder first transforms the time-domain filter coefficients into the concatenation of the real and imaginary components of the Real Fast Fourier Transform (RFFT) of the filter coefficients, then feeds this representation into the following fully connected model.
\begin{equation} \label{enc}
\begin{aligned}
    \bar{\mathbf{w}} &= [\Re (\text{RFFT}(\mathbf{w})), \Im (\text{RFFT}(\mathbf{w}))]\\
    \mathbf{z} &= \mathcal{E}(\mathbf{w})=(\mathcal{E}_2 \circ \sigma \circ f_\text{layernorm} \circ \mathcal{E}_1) (\bar{\mathbf{w}}),
\end{aligned}
\end{equation}
where $\mathcal{E}_1, \mathcal{E}_2$ are fully connected layers, $z$ is that latent variable that gets adapted in LFxLMS, $w$ is the time-domain representation of the filter coefficients, $\circ$ is the composition map, $ f_\text{layernorm}$ is the layer normalization,  $\sigma$ is the Sigmoid Linear Unit (SiLU) as the activation function, and $\Re, \Im$ represents the real and imaginary components of a complex number. 

The decoder takes a similar structure using the fully connected layers $\mathcal{D}_1, \mathcal{D}_2$ and transforms their output to time-domain filter coefficients.
\begin{equation}
\begin{aligned}
    \left [ \bar{\mathbf{w}}_1, \bar{\mathbf{w}}_2 \right ] &=(\mathcal{D}_2 \circ \sigma \circ f_\text{layernorm} \circ \mathcal{D}_1)(\mathbf{z}) \\
    \hat{\mathbf{w}} &= \mathcal{D}(\mathbf{z}) = \text{IRFFT }(\bar{\mathbf{w}}_1 + j \bar{\mathbf{w}}_2),
\end{aligned}
\end{equation}
where $j$ is the imaginary constant. All the models are designed using PyTorch, which allows for the transformation between filter representations using automatic differentiation \cite{autograd}. 
\subsection{Variational Auto-Encoders}
The Variational Auto-Encoder (VAE) aims to produce a latent space distributed as a standardized Gaussian \cite{vae_paper}. This allows the latent variables to be uncorrelated (also known as disentangled) and the decoder to be used as a generator of input samples. The encoder of the VAE $\mathcal{E}_{\text{VAE}}$ is slightly different from that of Equation (\ref{enc}) because the layer $\mathcal{E}_2$ outputs statistical parameters from which the latent variable is sampled. Using an auto-encoder model with a disentangled latent space could benefit latent adaptation because otherwise the latent gradient vector could point orthogonally to inherent correlations between latent variables.


 However, in \cite{infovae}, they demonstrate that maximizing the VAE ELBO can sometimes lead to a posterior mode collapse, where the decoder and encoder of the VAE fail to output a diverse set of samples. The Information Variational Auto-encoder (InfoVAE) \cite{infovae} addresses the issue of posterior mode collapse by adding the mutual information term $ I(x;z)$ in the objective. This leads to replacing the K-L divergence loss of a VAE \cite{vae_paper} with the Maximum-Mean Discrepancy \cite{mmd}, which is a kernel-based loss. This paper implements the \cite{infovae} model with parameters $\alpha =1, \lambda = 1000$.

\subsection{Auto-encoder with uniform mixup}
\begin{figure}[ht]
 \centerline{
 \includegraphics[width=\columnwidth]{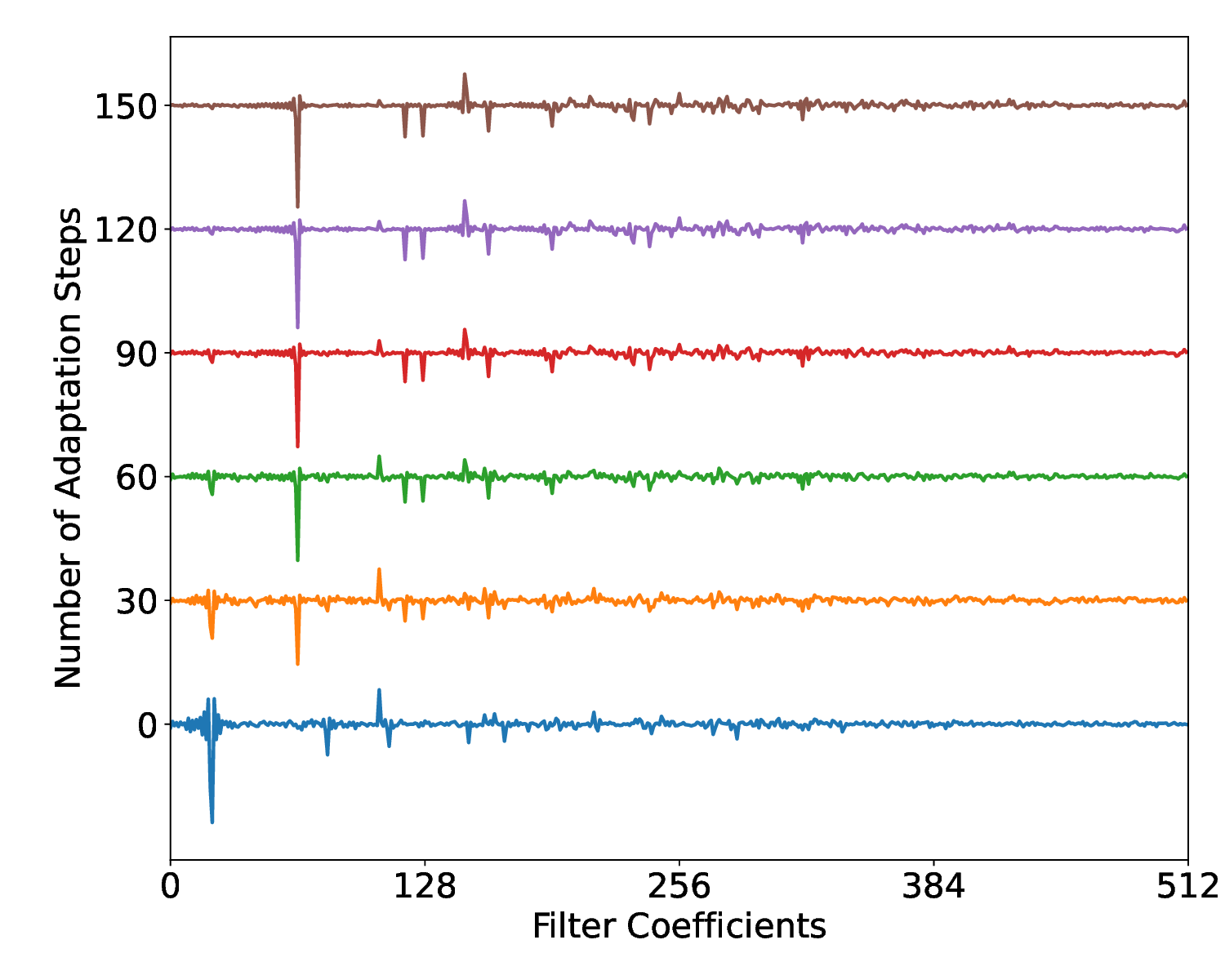}}
 
 \caption{ The FxLMS adaptive filter weights as the source position moves. Notice that the filter progressively removes the initial values of the filter while adding in the filter's final values. }
 \label{fig:filt_tracking}
\end{figure}
When the noise source location changes, the FxLMS algorithm will adapt $w$ to account for this change, removing components of the filter values at the beginning while adding components of the filter values at the end, as shown in Figure \ref{fig:filt_tracking}. The intermediate stages of the adaption process will likely not correspond to the space of converged $w$ filters that would normally be used to train the auto-encoder. Thus, we examine what happens when the auto-encoder is trained on convex combinations of random data pairs. This gives the following loss, to be added to the reconstruction loss of the auto-encoder
\begin{equation}
\begin{aligned}
    \mathcal{L}_C = E_{\mathbf{w}_1, \mathbf{w}_2 \in \mathcal{W}, \gamma \in U[0,1]} [(\gamma \mathbf{w}_1 + (1-\gamma) \mathbf{w}_2 \\
    - \mathcal{D}\circ \mathcal{E} (\gamma \mathbf{w}_1 + (1-\gamma) \mathbf{w}_2))^2],
\end{aligned}
\end{equation}
where $U[0,1]$ is a uniform distribution. On top of this, we also enforce that the latent variables match the convex combination of samples of $\mathcal{W}$ with the following loss 
\begin{equation} \label{eq:simp}
\begin{aligned}
    L_z &=  E_{\mathbf{w}_1, \mathbf{w}_2 \in \mathcal{W}, \gamma \in U[0,1]} [ \gamma \mathcal{E}( \mathbf{w}_1 )\\
    &+ (1-\gamma) \mathcal{E}(\mathbf{w}_2) - \mathcal{E}(\gamma \mathbf{w}_1 + (1-\gamma) \mathbf{w}_2)]^2,
\end{aligned}
\end{equation}

This expectation is approximated within each training batch of $\mathcal{W}$. Note that when a Dirichlet distribution instead models $\gamma$, the combined loss is a technique called 'mixup' \cite{mixup}. In mixup, this process is used for data augmentation, i.e., to generate synthetic samples to produce a richer mapping than what the dataset provides. 



%% file: text/Experiments.tex
\section{Experiments}\label{sec:experiments}
\begin{figure}[ht]
 \centerline{
 \includegraphics[width=0.6\columnwidth]{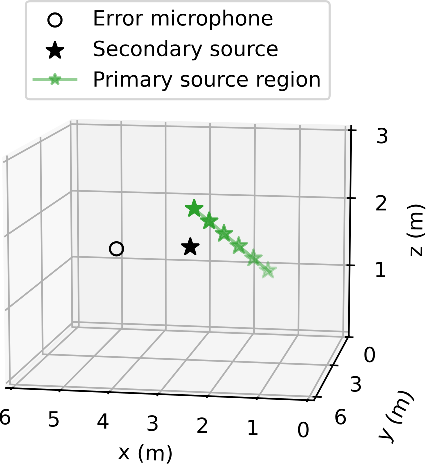}}
 
 \caption{ A room layout with the experimental parameters. The source positions are constrained to a line as the LFxLMS algorithm cancels the sound at the error microphone.}
 \label{fig:room_geo}
\end{figure}
We compare the performance of FxLMS with the LFxLMS model under various neural network constraints in a simulated acoustic environment. This experiment focuses on three considerations for LFxLMS: the impact of the VAE and infoVAE decoder, the improvement of convergence when the decoder is trained with the mixup constraint, and a comparison of the two normalized updates in the latent space. Currently, it is difficult to theoretically assess latent adaptive filtering. However, empirical methods can still yield valuable insights. This paper shows similar plots of ANC error signals and provides empirical measurements of convergence times.


\subsection{Simulated ANC with the Image Source Model}
We simulated ANC in a reverberant room using \textit{pyroomacoustics} \cite{PyRoom}, a Python implementation of the Image Source Model (ISM) \cite{allen}. In a $6 \times 6.2 \times  3$ m room with 0.15 s RT60 reverberation time, a secondary speaker is placed at the coordinates $[3, 2.5, 1.5]$ m, and an error microphone is placed at $[4.5, 3, 1.5]$ m. The region of primary source location is a line segment from $[1.5,1,1]$ m to $[3,2,2]$ m, of which we uniformly sample 2,048 primary source positions. Note that each primary source position has a roughly 2 mm spacing and that the primary source is farther away from the error microphone than the secondary source at every point. We obtain impulse responses of length $512$ samples from the secondary and primary paths at a sampling rate of $16,000$ Hz. The FxLMS algorithm is performed at each primary source position, and we obtain the converged filters to form $\mathcal{W}$.

From there, we train various auto-encoders with different constraints on $\mathcal{W}$. As mentioned in Section \ref{sec:neural}, we use a two-layer network for the encoder and decoder across all examined models. The hidden size was $256$, and the latent dimension was $32$.  The mixup constraint was implemented with $256$ convex combinations per batch with a batch size of $64$. The InfoVAE constraint is run with using a Gaussian kernel with kernel variance $0.01$.

The FxLMS experiments are performed in blocks of $100$ samples corresponding to 6.25 ms, with the weights starting from the $0$ vector.  The maximum stable step sizes must be found for each LFxLMS model to examine convergence speed across models. Determining the upper limits of the latent step-size for LFxLMS is difficult, so the maximum stable step size was determined empirically for each model. The latent-normalized models tend to require step-sizes within $[.05, .3]$, and the data-normalized models used step-sizes within $[5, 100]$. The primary source position is randomly initialized for each trial and switches to another random position at the 100th block to see how each model adapts to the rapid change. The error signals are collected and averaged from $50$ trials per model, which is then analyzed.

\subsection{Results on Simulated Data}
\begin{figure}[ht] 
 \centerline{
 \includegraphics[trim={0 1cm 0 0},clip,width=7.8cm]{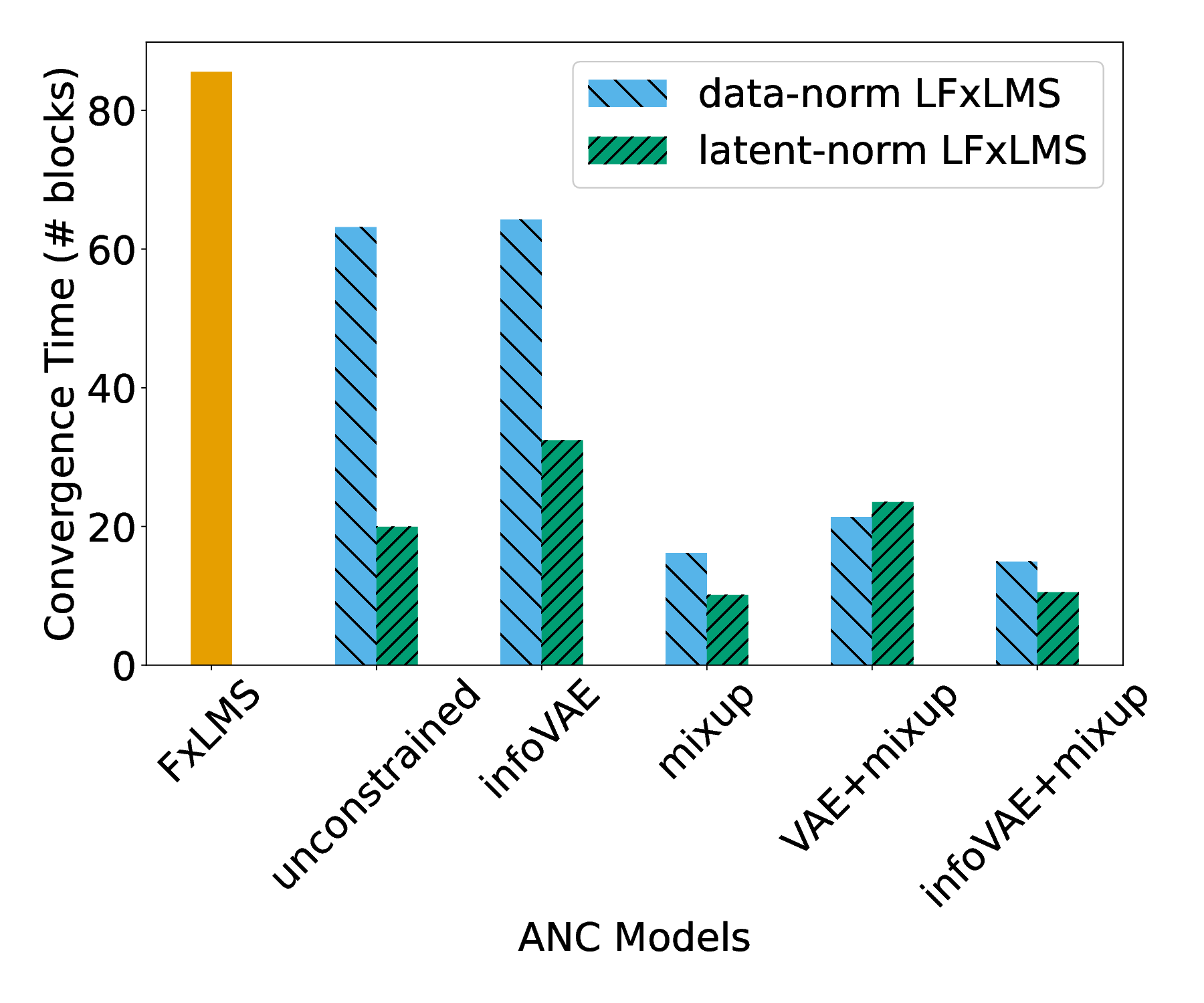}}
 \caption{The convergence time of FxLMS and LFxLMS under various neural network constraints and update normalization schemes measured in blocks of samples. Data-norm stands for data-normalization, and lat-norm stands for latent-normalization. The convergence time of LFxLMS with the VAE decoder is not shown.}
 \label{fig:result_convergence}
\end{figure}

This paper uses empirical methods of measuring the convergence rate of the LFxLMS algorithms. Section 23A of \cite{Sayed} gives the following empirical measure of convergence,
\begin{equation} \label{conv}
    C(e) = \arg \min_{k} e_k^2 \leq (1+\rho) \; e^2_\infty,
\end{equation}
where $e$ is the error signal, and $\rho$ is a small positive constant. If $\rho = 0.1$, it means the earliest sample in which the squared error is within $10\%$ of the steady-state mean squared error. 

Figure \ref{fig:result_convergence} shows the convergence times of the various LFxLMS models compared to FxLMS. For each model, the convergence times were estimated with $\rho = 0.4$ and averaged over the $50$ trials. To obtain an estimation of $e_\infty^2$, the error signal was averaged over the last $40$ blocks.

\begin{figure}[ht] 
 \centerline{
 \includegraphics[trim={0 1.3cm 0 0},clip,width=\columnwidth]{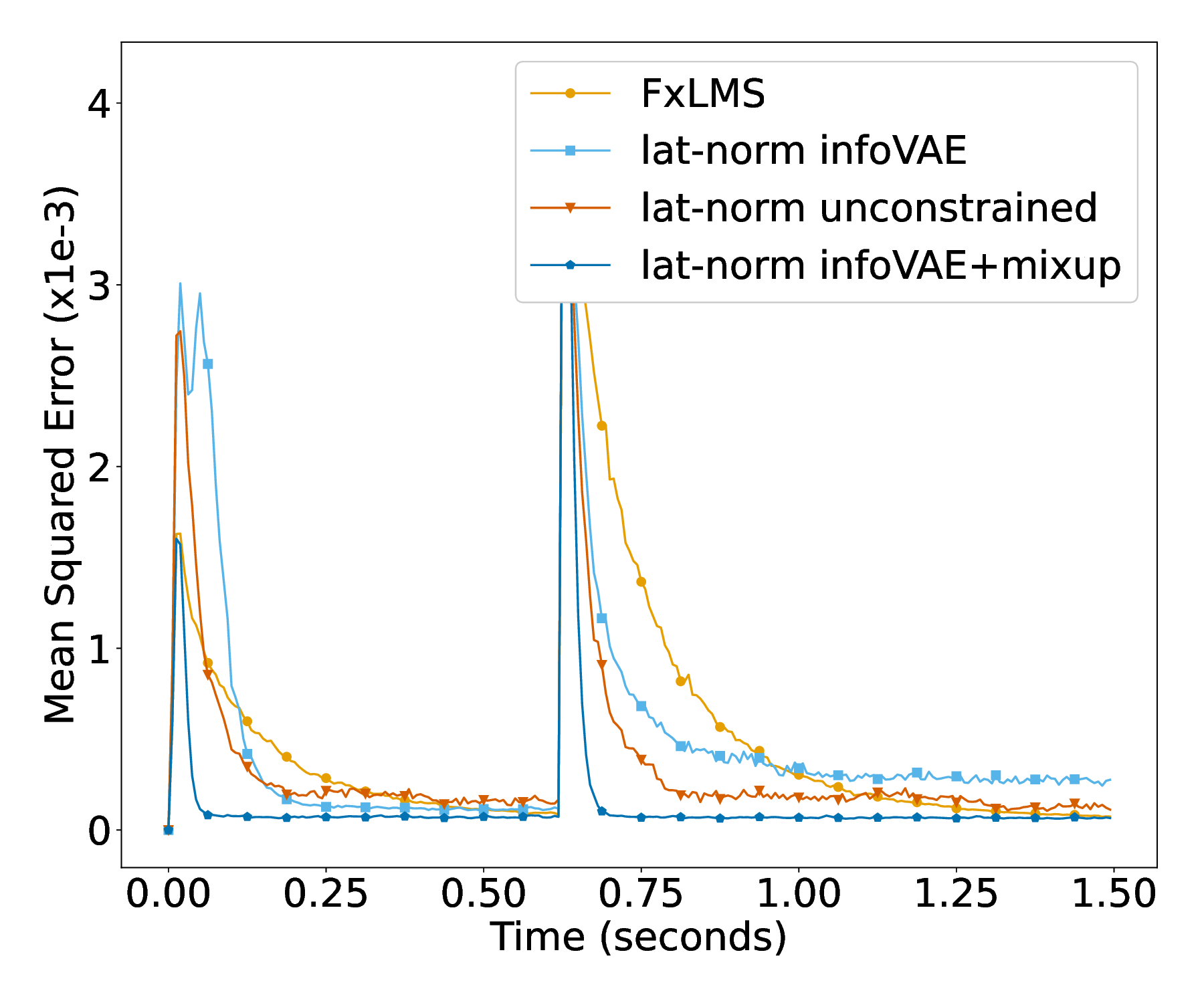}}
 \caption{ The averaged error signal is taken from latent-normalized models. The latent-normalized mixup LFxLMS is not shown because it is visually indistinguishable from the mixup-constrained infoVAE model.}
 \label{fig:lat}
\end{figure}

For all LFxLMS models shown, the convergence time is faster than the baseline. For most of the models, the latent-normalization leads to faster convergence. As for neural constraints, the models with the mixup constraint lead to faster convergence compared to models without this constraint. The LFxLMS model with the shortest convergence time is the latent-normalized LFxLMS with an InfoVAE decoder trained with mixup. Figure \ref{fig:lat} shows the averaged error signal, which supports the results from Figure \ref{fig:result_convergence}.

\begin{figure}[ht] 
 \centerline{
 \includegraphics[trim={0 1.3cm 0 0},clip,width=\columnwidth]{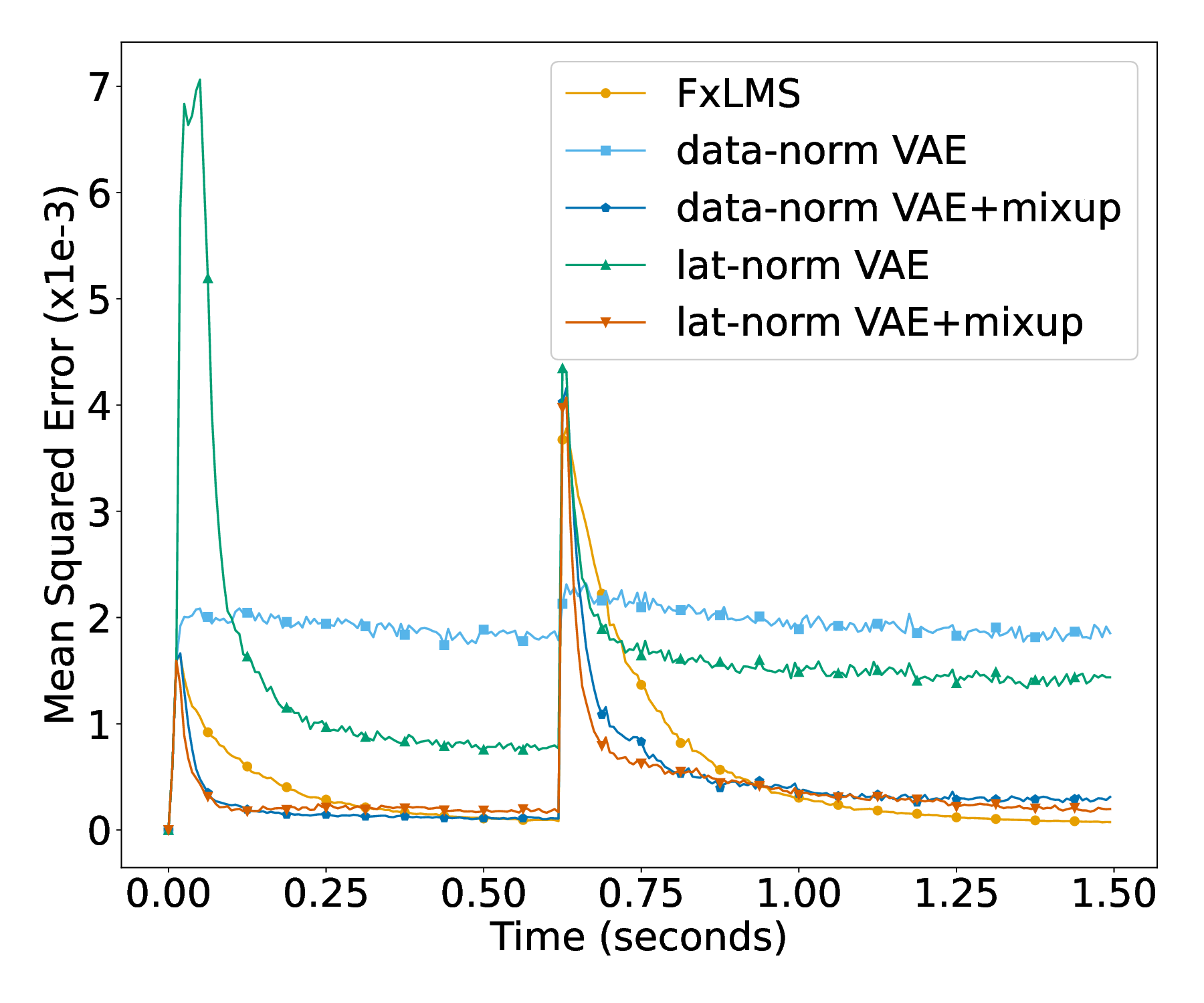}}
 \caption{ The averaged error signal is taken from all standard VAE models compared to FxLMS.}
 \label{fig:VAE}
\end{figure}
The steady-state ANC performance per model and trial is calculated as
\begin{equation}
    10 \log \frac{e_{\infty, \text{ANC OFF}}^2}{e_{\infty, \text{ANC ON}}^2},
\end{equation}
where $e^2_\infty$ was estimated using the last $40$ blocks of the error signal. Comparing across the models examined in Figure \ref{fig:result_convergence}, the ANC performance was averaged through the trials were all comparable. FxLMS had an average $37$dB drop in error energy, and all other models were within $3$dB of the FxLMS performance. This is surprising considering that the neural adaptive filters did not seem to follow the general tradeoff between convergence rate and steady state performance with traditional adaptive filters \cite{Sayed}.

The LFxLMS model that uses the VAE decoder was not included in Figure \ref{fig:result_convergence} because the VAE LFxLMS had a significantly worse convergence time and steady-state mean squared error than all other models. Figure \ref{fig:VAE} shows the averaged error signal of every VAE model compared to FxLMS. While the mixup loss combined with the VAE will still improve the overall ANC performance, Figure \ref{fig:result_convergence} shows better alternatives. 
%

%



%% file: text/Conclusion.tex
\section{Conclusion} \label{sec:conc}

We have found that performing the FxLMS update in the latent space of the decoder of an auto-encoder yields significant improvements in convergence speed. This is especially true when LFxLMS is performed with latent-normalized updates with a decoder trained with the mixup constraint. The improvements on ANC convergence using shallow neural networks show the potential promise of its capability for hardware integration with ANC devices. More work needs to be done on evaluating the interpolation capabilities of the LFxLMS algorithm. If an ANC application constrains the primary source position to a spatial region, the LFxLMS algorithm should be considered based on the work shown.

%% file: FA2025_template.bbl
\begin{thebibliography}{10}

\bibitem{survey1}
L.~Lu, K.-L. Yin, R.~C. {de Lamare}, Z.~Zheng, Y.~Yu, X.~Yang, and B.~Chen, ``{A survey on active noise control in the past decade—Part I: Linear systems},'' {\em Signal Processing}, vol.~183, p.~108039, 2021.

\bibitem{Riemann}
J.~Mejia, B.~Mederos, N.~Gordillo, and L.~Ortega, ``Adaptive filter with riemannian manifold constraint,'' {\em Scientific Reports}, vol.~13, no.~1, p.~9014, 2023.

\bibitem{Karim}
K.~Helwani, P.~Smaragdis, and M.~M. Goodwin, ``{Generative Modeling Based Manifold Learning for Adaptive Filtering Guidance},'' in {\em ICASSP 2023 - 2023 IEEE International Conference on Acoustics, Speech and Signal Processing (ICASSP)}, pp.~1--5, 2023.

\bibitem{EKF}
T.~Hardenbicker and P.~Jax, ``{Online System Identification on Learned Acoustic Manifolds Using an Extended Kalman Filter},'' in {\em 2024 18th International Workshop on Acoustic Signal Enhancement (IWAENC)}, pp.~339--343, 2024.

\bibitem{metaAF}
J.~Casebeer, N.~J. Bryan, and P.~Smaragdis, ``{Meta-AF: Meta-Learning for Adaptive Filters},'' 2022.

\bibitem{naf}
A.~Luo, Y.~Du, M.~J. Tarr, J.~B. Tenenbaum, A.~Torralba, and C.~Gan, ``{Learning Neural Acoustic Fields},'' 2023.

\bibitem{brendel}
A.~Brendel, J.~Zeitler, and W.~Kellermann, ``Manifold learning-supported estimation of relative transfer functions for spatial filtering,'' 2021.

\bibitem{vae_paper}
D.~P. Kingma and M.~Welling, ``{An Introduction to Variational Autoencoders},'' {\em Foundations and Trends® in Machine Learning}, vol.~12, no.~4, p.~307–392, 2019.

\bibitem{infovae}
S.~Zhao, J.~Song, and S.~Ermon, ``{InfoVAE: Information Maximizing Variational Autoencoders},'' 2018.

\bibitem{mixup}
H.~Zhang, M.~Cisse, Y.~N. Dauphin, and D.~Lopez-Paz, ``{mixup: Beyond Empirical Risk Minimization},'' in {\em International Conference on Learning Representations}, 2018.

\bibitem{survey2}
L.~Lu, K.-L. Yin, R.~C. {de Lamare}, Z.~Zheng, Y.~Yu, X.~Yang, and B.~Chen, ``{A survey on active noise control in the past decade–Part II: Nonlinear systems},'' {\em Signal Processing}, vol.~181, p.~107929, 2021.

\bibitem{flann}
G.~L. Sicuranza and A.~Carini, ``{A Generalized FLANN Filter for Nonlinear Active Noise Control},'' {\em IEEE Transactions on Audio, Speech, and Language Processing}, vol.~19, no.~8, pp.~2412--2417, 2011.

\bibitem{neuraliir}
C.-Y. Chang, ``{Neural filtered-U algorithm for the application of active noise control system with correction terms momentum},'' {\em Digital Signal Processing}, vol.~20, no.~4, pp.~1019--1026, 2010.

\bibitem{dlanc}
Y.-J. Cha, A.~Mostafavi, and S.~S. Benipal, ``{DNoiseNet: Deep learning-based feedback active noise control in various noisy environments},'' {\em Engineering Applications of Artificial Intelligence}, vol.~121, p.~105971, 2023.

\bibitem{rlanc}
Z.~Luo, H.~Ma, D.~Shi, and W.-S. Gan, ``{GFANC-RL: Reinforcement Learning-based Generative Fixed-filter Active Noise Control},'' {\em Neural Networks}, vol.~180, p.~106687, 2024.

\bibitem{rlanc2}
B.~Raeisy and S.~Golbahar~Haghighi, ``{Active Noise Controller with reinforcement learning},'' in {\em The 16th CSI International Symposium on Artificial Intelligence and Signal Processing (AISP 2012)}, pp.~074--079, 2012.

\bibitem{mcmlanc}
H.~Zhang and D.~Wang, ``{Deep MCANC: A deep learning approach to multi-channel active noise control},'' {\em Neural Networks}, vol.~158, pp.~318--327, 2023.

\bibitem{Sayed}
A.~H. Sayed, {\em {Adaptive Filters}}.
\newblock Hoboken, NJ: Wiley-Blackwell, 2008.

\bibitem{autograd}
A.~Paszke, S.~Gross, S.~Chintala, G.~Chanan, E.~Yang, Z.~DeVito, Z.~Lin, A.~Desmaison, L.~Antiga, and A.~Lerer, ``Automatic differentiation in pytorch,'' 2017.

\bibitem{mmd}
A.~Gretton, K.~Borgwardt, M.~J. Rasch, B.~Scholkopf, and A.~J. Smola, ``{A Kernel Method for the Two-Sample Problem},'' 2008.

\bibitem{PyRoom}
R.~Scheibler, E.~Bezzam, and I.~Dokmanic, ``{Pyroomacoustics: A Python Package for Audio Room Simulation and Array Processing Algorithms},'' in {\em 2018 IEEE International Conference on Acoustics, Speech and Signal Processing (ICASSP)}, p.~351–355, IEEE, Apr. 2018.

\bibitem{allen}
J.~Allen and D.~Berkley, ``{Image method for efficiently simulating small-room acoustics},'' {\em The Journal of the Acoustical Society of America}, vol.~65, pp.~943--950, 04 1979.

\end{thebibliography}
